\documentclass[a4paper]{article}

\usepackage{INTERSPEECH2021}

\usepackage{amsmath,graphicx}
\usepackage{xcolor}
\usepackage{comment}
\usepackage{hyperref}
\usepackage{tabularx}
\usepackage{scrextend}

\title{SwissDial: Parallel Multidialectal Corpus of Spoken Swiss German}
\name{Pelin Dogan-Sch\"{o}nberger, ~~Julian M\"{a}der, ~~Thomas Hofmann}

\address{ETH Z\"{u}rich, Department of Computer Science, Switzerland}
\email{\{pelin.dogan, maederju, thomas.hofmann\}@inf.ethz.ch}

\begin{document}

\maketitle
\begin{abstract}
Swiss German is a dialect continuum whose natively acquired dialects significantly differ from the formal variety of the language. These dialects are mostly used for verbal communication and do not have standard orthography. This has led to a lack of annotated datasets, rendering the use of many NLP methods infeasible. In this paper, we introduce the first annotated parallel corpus of spoken Swiss German across 8 major dialects, plus a Standard German reference.
Our goal has been to create and to make available a basic dataset for employing data-driven NLP applications in Swiss German. We present our data collection procedure in detail and validate the quality of our corpus by conducting experiments with the recent neural models for speech synthesis.
\end{abstract}


\section{Introduction}
\label{sec:intro}
In recent years, research in natural language processing has led to significant advancements on tasks such as sentiment analysis, question-answering, text- and speech-based machine translation, automatic speech recognition, or speech synthesis.  Intelligent language systems such as voice assistants, chat bots, or AR/VR devices start to permeate many aspects of our daily lives. 
These novel forms of human-computer interaction require suitable content and crucially depend on the availability of powerful models along with adequate data to train them. This creates inequalities across different languages and particularly poses challenges for low-resource languages and dialects, where data of sufficient quantity and quality is often scarce.

Among the low-resources languages, Swiss German dialects are characterized by being derived from Standard German (High German, DE) with considerable differences in phonetics, vocabulary, morphology and syntax, even lacking clear boundaries and definitions. 
Unlike other languages, where dialect usage decreases in favor of standard variants in most social domains, Swiss German dialects are widely used in everyday life (cf.~\cite{liidi2007swiss}). With the expansion of personalized digital media and social platforms, dialect use is growing even more. Despite this increasing demand, the obstacles posed by, for instance, lack of standard orthography, have made data collection of written text (e.g.~transcripts) challenging. These difficulties in collecting consistent and clean data have led to a situation, where Switzerland is lagging behind and struggles to take full advantage of basic NLP tools.

\begin{figure}[t]
\centering
\includegraphics[trim=65 0 350 0, clip, width=0.5\textwidth]{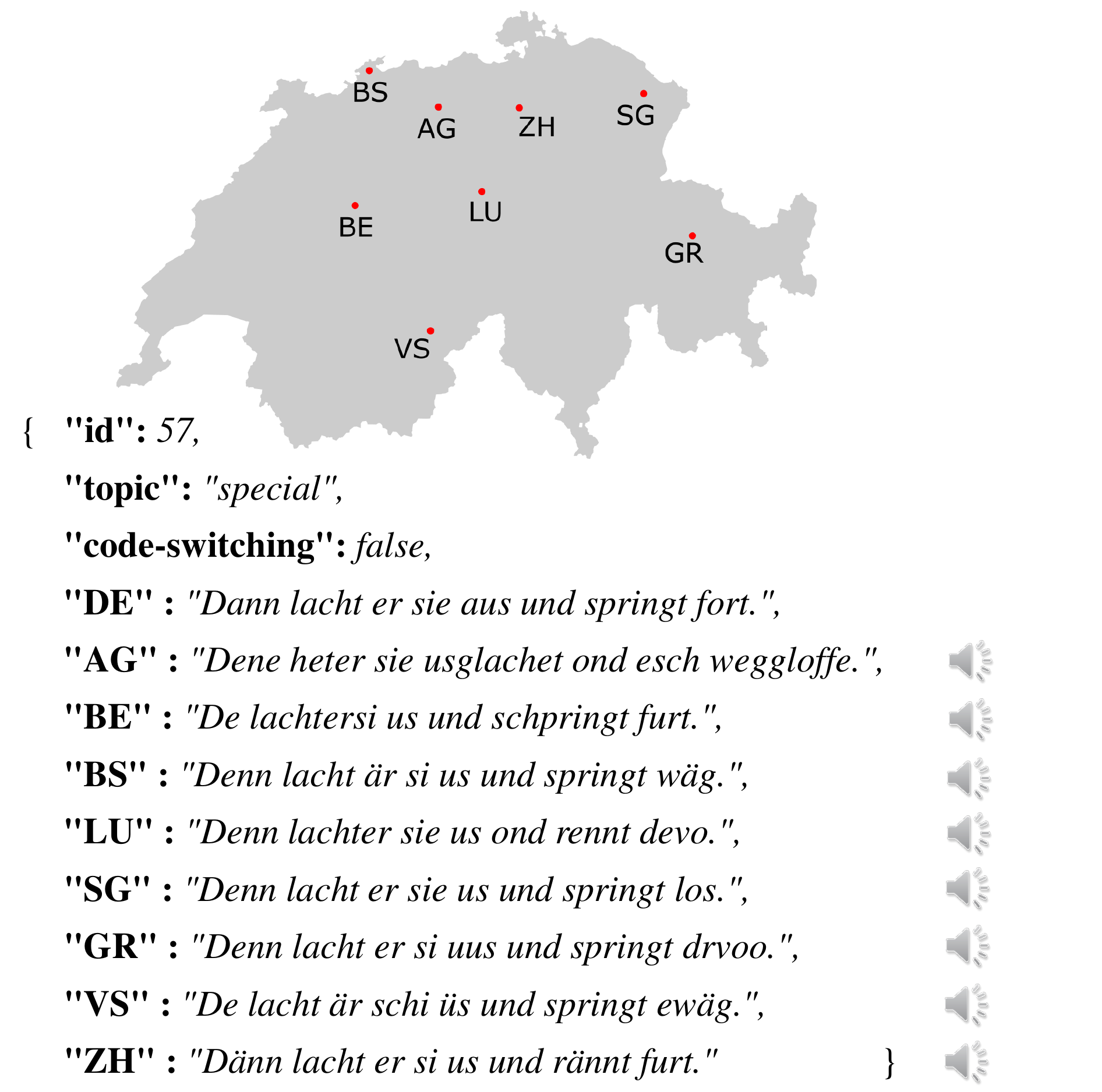}
\caption{\textit{A sample sentence from SwissDial. For each sample sentence in High German, we provide the manual translations and the speech samples in 8 Swiss dialects. (The sentence means \textit{"Then he laughs at her and jumps away."} in English)}}
\label{fig:sample}
\end{figure}

This paper presents an annotated parallel corpus of spoken Swiss German dialects, \emph{SwissDial}, for 8 different regions: Aargau (AG), Bern (BE), Basel (BS), Graub\"{u}nden (GR), Luzern (LU), St. Gallen (SG), Wallis (VS) and Z\"{u}rich (ZH). Our corpus consists of web-crawled sentences in High German, manual translations into Swiss German dialects and their audio recordings. Different genres and sources are represented: news stories, Wikipedia articles, weather reports, short stories. The audio recordings are performed by a single speaker for each dialect. Figure \ref{fig:sample} presents a sample from SwissDial. Our primary motivation and focus is to provide a basic dataset that would allow applicability of recent data-driven NLP models, especially for speech synthesis, and step up the ongoing research in Swiss German. To our knowledge, this is the  first publicly available parallel corpus of spoken Swiss German. 
Our main contributions are as follows:
\begin{itemize}
    \item We present \textit{SwissDial}, a parallel corpus of text and audio in 8 Swiss German dialects on designated topics. The data collection process is well-documented, allowing for transparency and augmentability. The corpus is available at \url{https://projects.mtc.ethz.ch/swiss-voice-data-collection}.
    \item We validate the suitability of our corpus by running experiments with the recent neural models for (i) single speaker, (ii) multi-speaker-multi-dialect data, and (iii) code-switching (CS) speech synthesis. The samples from these models are available at \url{https://projects.mtc.ethz.ch/projects/swiss-voice/swissdial}.
\end{itemize}

\section{Related Work}
For various high-resource languages, there are parallel text corpora \cite{jawaid2014tagged, koehn2005europarl, buck2014n} and monolingual text corpora such as \cite{kupietz2010german}, Google News, Common Crawl that are used for the tasks of language modeling and machine translation. \cite{panayotov2015librispeech, ljspeech17, park2019css10} present paired audio and text data which are widely used for automatic speech recognition and speech synthesis with various transformations. For these languages, \cite{di2019must} presents a parallel corpus for multilingual speech translation. 

However, availability of data is limited for languages containing strong dialectal variations such as Arabic, Chinese, Swiss German which come forward with their special condition where the formal variety of the language differs significantly from varieties that are acquired natively and mostly used for verbal communication.
There are various datasets concerning Swiss German that allow various NLP tasks such as dialect identification \cite{ali2018character, jauhiainen2018heli}, dialect machine translation \cite{honnet2018machine, scherrer2016automatic}, morphology generation \cite{scherrer2011morphology}, automatic speech recognition~\cite{garner2014automatic}. 
Archimob \cite{samardzic2016archimob, scherrer2019digitising} presents a general purpose corpus of spoken Swiss German based on oral history interviews with various people from different dialects, which provides pairs of audio and text with annotated normalization and part-of-speech tagging. 
Another resource~\cite{swiss_sms}, Swiss SMS corpus, provides text corpus of SMS messages. 
NOAH's corpus  \cite{hollenstein2014compilation} presents a collection of text in various genres in Swiss dialects with manually annotated part-of-speech tags. 
\cite{honnet2018machine} collects a various parallel written sources (High German and Swiss German) to perform machine translation by normalizing Swiss German. 
Recently, \cite{schmidt2020swiss} introduced a dictionary containing forms of common words in various Swiss German dialects normalized into High German. 

Most of these existing resources in Swiss German do not provide sufficient data to train end-to-end models for neural machine translation and speech synthesis. They rather provide word-level correspondences which are suboptimal for sentence level NLP due to the lack of full context and syntax. Moreover, existing datasets of spoken language do not fulfill the requirements of today's neural speech synthesizers, which ideally need clean and single speaker audio. SwissDial is thus, to our knowledge, the first available parallel multi-dialectal corpus with transcribed text and clean audio in Swiss German. The text part of our corpus is carefully designed to cover a large set of topics and lexicons providing manual and parallel translations of sentences in dialects. The audio part, paired with the corresponding text, provides clean and single speaker audio for each dialect at sentence level. Moreover, the corpus contains labeled code-mixing samples which would be valuable for code-mixing speech synthesis research. We believe that the parallel structure of our corpus can also be used to improve data efficiency in NLP tasks via transfer learning. 

\begin{table*}[t]
\centering
\begin{tabular}{lrrrrrrrrrrr}
\hline \textbf{Topic} &  \textbf{Source} & \textbf{Parallel} & ~\textbf{AG} & \textbf{BE} & \textbf{BS} & \textbf{GR} & \textbf{LU} & \textbf{SG} & \textbf{VS} & \textbf{ZH} & \textbf{Total} \\ \hline
Animals/farming & News & 130 & 137 & 135 & 137 & 137 & 135 & 137 & 137 & 134 & 1'089 \\
Culture & News  & 81 & 85 & 87 & 87 & 86 & 86 & 87 & 87 & 86 & 691 \\
Earth/Space & Wikipedia  & 17 & 18 & 18 & 18 & 18 & 18 & 18 & 18 & 25 & 151 \\
Economics & News  & 60 & 63 & 63 & 62 & 63 & 63 & 63 & 63 & 234 & 674 \\
Int. politics & News  & 140 & 155 & 154 & 153 & 155 & 152 & 155 & 155 & 151 & 1'230 \\
Medicine & News  & 104 & 111 & 110 & 109 & 112 & 110 & 112 & 112 & 140 & 916 \\
Meteorology & News  & 192 & 215 & 201 & 214 & 215 & 202 & 215 & 215 & 211 & 1'688 \\
Random & Wikipedia  & 132 & 149 & 142 & 142 & 148 & 148 & 149 & 147 & 160 & 1'185 \\
Special & Wikipedia  & 1030 & 1'120 & 1'105 & 1'106 & 1'120 & 1'112 & 1'122 & 1'123 & 1'730 & 9'538 \\
Science & News & 189 & 197 & 192 & 195 & 197 & 194 & 197 & 197 & 243 & 1'612 \\
Sports & News  & 88 & 103 & 101 & 101 & 103 & 97 & 102 & 103 & 163 & 873 \\
Story & Internet  & 49 & 50 & 50 & 50 & 49 & 50 & 50 & 50 & 81 & 430 \\
Swiss politics & News  & 228 & 248 & 245 & 245 & 249 & 241 & 249 & 249 & 240 & 1'966 \\
Swiss regional & News  & 88 & 97 & 97 & 94 & 97 & 97 & 96 & 97 & 467 & 1'142 \\
\hline
with Code-Switching & News  & 232 & 247 & 247 & 247 & 249 & 248 & 250 & 248 & 345 & 2'081 \\
\hline
All topics & - & 2528  & 2'748 & 2'700 & 2'713 & 2'749 & 2'715 & 2'752 & 2'753 & 4'065 & 23'195 \\
\hline
\hline
\end{tabular}
\caption{\label{table:topic_distribution} The distribution of sentences among dialects. \emph{Special} indicates the sentences that are manually prepared to catch different vocabulary among the dialects.}
\end{table*}

\section{SwissDial Corpus Content and Creation}

In the following, we present the construction and content of our corpus and its modalities. Acquisition of SwissDial for applications in end-to-end neural models has five main steps: obtaining raw data, selection of dialects and annotators, translation, recording, and post-processing.

\subsection{Raw data}
\label{section:raw_data}
Swiss dialects are spoken colloquially, but only rarely used in written form. Moreover, the available written forms collected from unspecific sources generally contain inconsistencies within the same dialect, since there is no official standard orthography. This poses a challenge for annotators, possibly leading to disagreements with regard to the orthography of the provided text. To increase consistency and reduce ambiguity of the process, we started by collecting High German sentences to be translated by the annotators themselves.

We collected random sentences from news articles to cover a wide range of topics for good generalization. These sentences were labeled with the topic metadata of the article that they belong to. As mentioned earlier, Swiss dialects differ in vocabulary from High German and from each other. We thus crawled the internet for lists of lexical items that show variability among the dialects. We then extracted sentences from Wikipedia articles that contain each word (labeled as \textit{special}). In this way, we aimed to cover vocabulary differences of dialects, while having a wide range of topics for a good generalization in various NLP tasks. Table~\ref{table:topic_distribution} presents the topic distribution of the collected sentences in the corpus. 
It is important to note that significant majority of the code-switching (CS) content was the result of the random selection procedure, rather manual addition. There are two main reasons for this relatively large amount of CS content: (i) there is a high linguistic contact between English and German which results in considerable amount of word \textit{borrowing}, (ii) our main sentence source is news articles that contain global topics like politics, science, technology etc, and therefore naturally contain words from English.

\subsection{Selection of dialects}
Swiss dialects are not clearly separated by geological borders, the variation in the regional language is rather continuous. One can separate each dialect into numerous sub-dialects even down to resolution of villages.
Therefore, we had to make discretization where we aimed to represent the differentiation of the dialects as much as possible while addressing the most populated areas. Our selection motivation was to cover a range of dialects while complying with time constraints and availability of annotators. 
As a result, our corpus contains data from the following regions: Aargau, Bern, Basel, Graub\"{u}nden, Luzern, St. Gallen, Wallis and Z\"{u}rich.
To represent these areas, we carried out an auditioning where we evaluated the annotator candidates based on their performance on a sample translation and pronunciation task, as well as their vocal quality features inspired by \cite{kempster2009consensus}. We hired one non-professional annotator for each dialect that passed all the tests to perform the translation and audio recording steps.

\subsection{Text translation}
We asked annotators from each dialect to translate the provided High German sentences, described in Section \ref{section:raw_data}, to their native dialect. As mentioned earlier, Swiss dialects have different morphology together with different vocabulary and word choice. In some cases, a High German sentence or expression may not be easily translatable into dialect without sounding artificial. Therefore  annotators had the freedom to skip such sentences. All the  numbers, abbreviations and time expressions were translated as the annotator would read them out loud in the next step. In other words, the translation step aimed to have a one-to-one mapping between the text and the audio to-be-recorded. The foreign words, e.g. code-switching words, in the High German sentences were directly transferred as in the original language, rather than performing translation or phone mapping.

\subsection{Recording}
As the last step of the data collection, the annotators were asked to read out loud their translations in a quiet room, in front of a high-quality recording set-up to obtain clean recordings. We used a Neumann TLM 102 microphone with pop filter and recorded in mono at 48 kHz. The speakers were instructed to speak at a distance of about 15 cm from the microphone. We prepared a user interface that shows the annotator the translated sentences each at a time and allows recording control through keyboard. In this way, the annotators were able to prepare and control their voice and breathing for each sentence, and rest when they needed. The interface also allowed annotators to modify their translations when they notice typos or incorrect wording/expressions before recording each sentence. This can be considered as the last quality control step for the correctness of the translations. The recordings were carried out in multiple sessions, around the same time of the day, that took 1-2 hours each. This scheduling is important to have a consistent voice across all the audio samples without exhausting the annotators and their vocal chords.
Table~\ref{table:audio_statistics} presents the statistics of the collected text data.
\\
\begin{table}
\centering
\begin{tabular}{lccc}
\hline \textbf{Dialect} & \textbf{Total} & \textbf{Average} & \textbf{Average \#word}\\
  & \textbf{Audio (h)} & \textbf{utterance (s)} & \textbf{per utterance}\\ \hline
AG & 2.78 & 3.64 & 11.8\\
BE & 3.41 & 4.55 & 11.9 \\
BS & 3.15 & 4.18 & 12.7 \\
LU & 2.54 & 3.36 & 12.3 \\
GR & 2.93 & 3.83 & 12.5 \\
SG & 3.71 & 4.85 & 12.1 \\
VS & 3.32 & 4.34 & 11.9 \\
ZH & 4.55 & 4.03 & 12.2\\
\hline
\hline
\end{tabular}
\caption{\label{table:audio_statistics} The statistics of the collected audio data.}
\end{table}

\subsection{Post-processing}
After translations and recording, we performed small post-processing on the collected data to make it more convenient for various NLP tasks. For the text data in High German, we post-processed each sentence in the corpus by spelling numeric values into words. For the dialect text, we post-processed each sentence transforming the spelling of the numbers into numeric characters. In this way, we provide two versions for each sentence in all dialects: text with numeric characters and all characters. Providing both versions is helpful to tokenize numbers in various NLP tasks. 

For the audio recordings, we performed a cleaning procedure where the aim was to cut off the segments at the beginning and end of the utterances that contain various noises like key pressing, breathing, mouth noises, etc. Some of the utterances contained noise during the sentence reading, rather than beginning/end, which would not be straightforward to clean. We repeated the recording step for these samples whenever it was possible, and eliminated the rest to present clean data which is significant for the speech synthesis task. Then, we down-sampled all recordings to 22050 Hz to be used in our speech synthesis experiments.

\section{Speech Synthesis Experiments with SwissDial}
With the increase in monolingual corpora, the end-to-end architectures~\cite{shen2018natural, oord2016wavenet} have shown high quality speech synthesis results. Using these methods, we performed various experiments to synthesize speech in Swiss dialects using audio-text pairs from Swiss Dial. 
All the experiments below are carried out with character level input representations.

\begin{figure*}[t]
\centering
\includegraphics[trim=40 310 50 50, clip, width=0.9\textwidth]{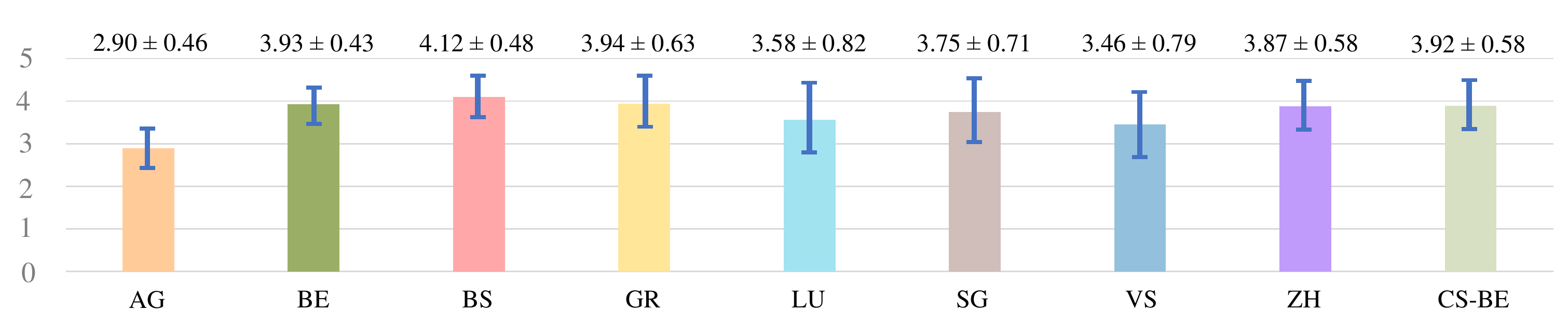}
\caption{\textit{MOS results on the single speaker model and code-switching model with standard deviation.}}
\label{fig:user_study}
\end{figure*}

\subsection{Single speaker model}
We implemented the well-established neural speech synthesis model \cite{shen2018natural}, an encoder-decoder architecture with attention, and conducted experiments on each dialect separately. 
To increase the data efficiency for our low-resource set-up, we firstly trained on the German part of CSS10~\cite{park2019css10} (16 hours) for a good initialization. The main motivation for pretraining on High German is to transfer the textual and acoustic representations and learned alignment from High German data to Swiss dialects in the tuning stage, since Swiss dialects are derived from High German. Then, we fine-tuned the pretrained model on each dialect separately to obtain a single speaker speech synthesis model. For each dialect, we held out 10\% of the samples as a validation set. 

We conducted a 5-scale mean opinion score (MOS) test, with 0.5 increments, for speech naturalness and quality. Fifteen native Swiss speakers from different regions were invited to participate. Randomly chosen ten utterances from a test set are presented to the participants for each dialect. Figure \ref{fig:user_study} shows the mean and deviation of MOS ratings of the participants. One has to keep in mind, that we hired non-professional speakers which naturally have, in comparison to professional speakers, much stronger deviations in quality and consistency of their pronunciation, intonation and voice control. Therefore, we suspect the differences in MOS rating between the dialects to be a result of the varying vocal quality of our non-professional speakers.

\subsection{Multi-speaker-multi-dialect model} 
Next, we aimed to encourage sharing of model capacity across different dialects by training a single model on all dialects simultaneously. We implemented the multi-speaker model in~\cite{jia2018transfer}, which is an extension of the previous model, and extended it with learnt dialect embeddings to support multi-dialect training. The speaker and the dialect embeddings are learnt in a similar way following~\cite{wan2018generalized}. 
We trained the speaker and dialect verification models by collecting samples from additional speakers from radio shows to enhance the capability of representations, since our dataset would only provide 8 different speakers. We held out 5\% of the samples as a validation set. Different from~\cite{jia2018transfer}, the learnt speaker and dialects embeddings are concatenated at the encoder output and the decoder input inspired by \cite{xue2019building}. An internal subjective comparison between the multi-speaker-multi-dialect model results and the single speaker model results revealed a very similar naturalness and quality for the two models.

\subsection{Code-Switching.} 
Lastly, we implemented the model~\textit{SE-DEC}~\cite{xue2019building}, which is again an extension on \cite{shen2018natural}, to explore the performance of our corpus in code-switching set-up with English words mixed in single Swiss dialect, BE, for showcasing. Firstly, we pretrained the model with monolingual recordings in English~\cite{ljspeech17} and High German~\cite{park2019css10}. Then we fine-tuned the model using monolingual English data and the BE part of our dataset which introduces 247 code-switching samples. The MOS result for this experiment is shown in Figure \ref{fig:user_study} with the CS-BE bar. The participants were asked to rate the samples according to (i) Swiss German pronunciation, (ii) English pronunciation, (iii) naturalness and quality of the audio. As one can see in Figure \ref{fig:user_study}, the resulting MOS ratings for BE with and without code-switching are almost identical, which is a very encouraging result for the use of code-switching extensions for Swiss German speech synthesis.

\section{Conclusion}
In this paper, we present SwissDial, a parallel multidialectal corpus of text and audio in 8 Swiss German dialects, by describing the collection procedure in detail. Furthermore, we validate the quality of SwissDial by presenting the results of experiments with the recent neural models for speech synthesis in single-speaker-single-dialect, multi-speaker-multi-dialect, and code-switching set-ups. Last but not least, we believe that applications on SwissDial are not limited only to speech synthesis but it can also empower further NLP research and application in Swiss German in the fields of machine translation and dialect identification.

\bibliographystyle{IEEEtran}

\bibliography{main}

\end{document}